\pdfoutput=1
\documentclass[11pt]{article}
\usepackage{nodalida2025}
\usepackage{times}
\usepackage{url}
\usepackage{latexsym}

\usepackage{graphicx}
\usepackage{tabularx}
\usepackage{multirow}
\usepackage[T1]{fontenc}
\usepackage{kotex}

\usepackage[utf8]{inputenc}
\usepackage{tikz-dependency}
\usepackage{forest}
\usepackage{microtype}

\usepackage{inconsolata}
\usepackage{hyperref}
\usepackage{graphicx}

\aclfinalcopy 

\title{Second language Korean Universal Dependency treebank v1.2: \\ Focus on data augmentation and annotation scheme refinement}

\author{Hakyung Sung \\
  Department of Linguistics \\
  University of Oregon \\
  {\tt hsung@uoregon.edu} \\\And
  Gyu-Ho Shin \\
  Department of Linguistics \\
  University of Illinois Chicago \\
  {\tt ghshin@uic.edu} \\}

\date{}

\begin{document}
\maketitle
\begin{abstract}
We expand the second language (L2) Korean Universal Dependencies (UD) treebank with 5,454 manually annotated sentences. The annotation guidelines are also revised to better align with the UD framework. Using this enhanced treebank, we fine-tune three Korean language models—Stanza, spaCy, and Trankit—and evaluate their performance on in-domain and out-of-domain L2-Korean datasets. The results show that fine-tuning significantly improves their performance across various metrics, thus highlighting the importance of using well-tailored L2 datasets for fine-tuning first-language-based, general-purpose language models for the morphosyntactic analysis of L2 data.
\end{abstract}

\section{Introduction}

The Universal Dependencies (UD) framework, designed to facilitate accessible morphosyntactic annotations \cite{de2021universal}, has been applied increasingly in linguistics, particularly to annotate learner corpora. This approach supports tasks such as modeling the trajectories of second language (L2) acquisition, which often require treebanks for fine-tuning language models or evaluating their performance on L2 data. Such data are typically characterized by simpler and/or nontarget-like lexico-grammatical usages compared to those produced by first-language speakers, although these characteristics vary across L2 proficiency. Previous research has increasingly adopted the UD framework to automatically handle learner corpora in various languages, including English \cite{berzak2016universal, kyle2022dependency, lyashevskaya2017realec, huang2018dependency}, Chinese \cite{lee2017towards}, Italian \cite{di2019towards, di2022valico}, Russian \cite{rozovskaya2024universal}, and Swedish \cite{masciolini2024synthetic, masciolini2023query, masciolini2023towards}, demonstrating its utility in L2 studies.

Among these efforts, recent studies in Korean have developed L2-Korean UD treebanks with language-specific morphemes and dependency tags \cite{sung2023diversifying, sung2023towards, sung2024constructing}. However, two research gaps remain. First, while continuing to expand the amount of data, annotation guidelines should be iteratively updated to balance cross-linguistic standardization with the preservation of language-specific features \cite{de2021universal, manning2011part}. Second, the effectiveness of L2-Korean-optimized models should be assessed using out-of-domain data to improve their reliability in broader contexts for which they are designed \cite{plank2016non, joshi2018extending}.

The present study addresses these gaps with three key contributions: (1) augmenting the existing L2-Korean UD treebank (v1.1, 7,530 sentences) by adding 5,454 manually annotated sentences with Korean-specific morphemes and UD annotations; (2) revising dependency annotation guidelines extensively to better align with the language-general UD framework, while implementing minor adjustments to the guidelines to better reflect the linguistic properties of Korean; and (3) fine-tuning and evaluating three Korean language models in both in-domain and out-of-domain contexts using the updated L2-Korean UD treebank (v1.2, 12,984 sentences, see Appendix for XPOS and DEPREL tag distributions).

\section{Related works}
A line of studies have established approaches for morpheme and dependency annotations in L2 Korean. \citet{sung2023towards} provided preliminary guidelines for Korean morpheme annotations, addressing the need to parse morphemes taking into account the agglutinative nature of Korean morphosyntax, where a single word often combines lexical morphemes (e.g., noun, verb) and functional morphemes (e.g., postpositions, tense-aspect-modality markers). Expanding this work, \citet{sung2024constructing} introduced detailed UD annotation guidelines to handle Korean-specific dependency cases such as particles and coordination.

\citet{sung2023diversifying} fine-tuned morpheme parsers optimized for L2 Korean and evaluated them on in-domain and out-of-domain datasets, demonstrating the importance of high-quality input for fine-tuning L2-Korean language models. However, those studies did not include training or evaluating dependency tags. Additionally, their fine-tuning strategy was relatively simple, relying solely on one Korean pre-trained model.

\section{Dataset}

Building upon the previous L2-Korean UD annotation projects \cite{sung2023towards, sung2024constructing}, we continued annotating L2-Korean sentences using a subset of data from the same source \cite{park2016korean}.\footnote{The source data became unavailable as of September 2024.} For the out-of-domain testing, we annotated additional data from the KoLLA dataset \cite{lee2022corpus}, which was designed to analyze Korean learner language with a focus on particle error annotations.\footnote{The dataset is publicly available at: \url{https://cl.indiana.edu/~kolla/}}

Along with the annotations, we refined the annotation guidelines, implementing major revisions to better align with the language-general UD annotation scheme and minor adjustments to morpheme annotations. Together, the updated L2-Korean UD treebank (v1.2) comprises (\# sents = 12,984): (1) additional data augmented and annotated using the revised scheme (\# sents = 4,532); (2) revised data from the previous project \cite{sung2024constructing}, updated with the new annotation scheme (\# sents = 7,530); (3) data sourced from the KoLLA dataset \cite{lee2022corpus}, annotated with the revised scheme for the out-of-domain testing (\# sents = 922).

\subsection{Refining annotation guidelines}

Carefully curated linguistic annotations balance two key challenges: maintaining consistency and ensuring accuracy. \citet{manning2011part} highlighted the challenges involving POS labeling, noting the inherent ambiguities and unclear boundaries between word classes, which complicate the definitive assignment of labels. Such intrinsic ambiguities can degrade the performance of taggers when training language models. Therefore, systematic checks and guideline refinements are essential for achieving optimal annotations.

For L2-Korean annotations, \citet{sung2024constructing} emphasized dependency annotations grounded in language-specific justifications, building upon earlier studies of Korean dependency annotations \cite{lee2019ko, Kim2018ko, seo2019korean}. However, the previous annotation scheme did not fully conform to the language-general UD framework and exhibited notable mismatches between tags, particularly \texttt{conj}, \texttt{flat}, and \texttt{aux}. To address these issues, we revised the previous dependency annotation guidelines to better align with the language-general UD conventions, thus enhancing global applicability. Below, we outline two key areas of major changes implemented.

\subsubsection{Following the left-to-right rule}
The UD framework enforces a strict left-to-right rule for coordination to ensure consistency and cross-linguistic applicability in morphosyntactic annotations \cite{nivre2016universal, de2021universal}. This approach originates from the Stanford-typed dependencies for English \cite{de2006generating}, which serve as the foundation for the universal dependency representation \cite{mcdonald2013universal}.

\paragraph{Coordination}
Coordination (\texttt{conj}) is handled by consistently attaching the coordinating conjunction to the head of the first conjunct. The leftmost conjunct is designated as the head, with subsequent conjuncts and the coordinating conjunction depending on it.\footnote{This approach, while widely adopted, has raised some questions, as noted by \citet{gerdes2016dependency}, where the selection of the first conjunct as the head is made without extensive justifications (p. 7).}

Initially, \citet{sung2024constructing} assigned the head to the right-headed structure in complex clauses or noun phrase conjunctions. For instance, in complex clauses, the head was assigned to the predicate, often resulting in a right-headed structure. This approach was driven by the nature of the Korean connective marker (e.g., -고 [\textit{-ko}]), which signifies conjunction and is logically tagged as \texttt{conj} (p. 3748). However, in line with the current UD guidelines, we revised the previous approach to strictly follow the left-to-right head structure, consistent with the UD’s left-headed coordination. Now, the connective marker -고 (\textit{-ko}) is tagged as \texttt{root}, and the final predicate receives the \texttt{conj} tag (Figure \ref{fig:1}).\footnote{We also revised noun phrase conjunctions, as in examples such as 사과와 바나나 (\textit{sakwa-wa} \textit{panana}, "apple and banana"), where 사과 (\textit{sakwa}, "apple") is the head and 바나나 (\textit{panana}, "banana") depends on it, with the coordinating conjunction -와 (\textit{-wa}, "and") linking the two.}

\begin{figure}
    \centering
    \resizebox{0.5\textwidth}{!}{
        \begin{dependency}
            \begin{deptext}
            저는 \& 구경을 \& \textbf{했고} \& 음식도 \& \textbf{먹었어요} \\
            \textit{ce-nun} \& \textit{kwukyeng-ul} \& \textit{hay-ss-ko} \& \textit{umsik-to} \& \textit{mek-ess-eyo}  \\
            I-TOP  \& sightseeing-ACC \& do-PST-CONN \& food-PAR \& eat-PST-DECL \\
            \end{deptext}
            \depedge{3}{1}{nsubj}  
            \depedge{3}{2}{obj}
            \deproot{3}{root}           
            \depedge{3}{5}{conj}
            \depedge{5}{4}{obj}
        \end{dependency}
    }
    \caption{Coordination (Left-headed) \\ \footnotesize{`I looked around and ate some foods.'}\\}
    \label{fig:1}
\end{figure}

\paragraph{Flat}
Flat (\texttt{flat}) is used when no single element in an expression can be clearly identified as the head. Similar to the case of coordination, in this structure, the leftmost element is treated as the head, with all subsequent components attached to it as equals. This applies to expressions such as "John Smith" or "San Francisco," where no one part dominates the meaning of the whole.

In the previous L2-Korean UD annotation scheme, the core principle for assigning the head was based on the presence of particles, reflecting how they function in determining the grammatical roles of nouns in Korean—core arguments (subject, object) or non-core arguments (obliques) within a clause \cite{sohn1999korean}. However, to conform to the UD framework's left-to-right rule, we rigorously revised all flat relations to follow this directionality. This revision affected the majority of naming conventions and combinations of names with titles in our annotated data, as described in Figure \ref{fig:2}.

\begin{figure}[h!]
    \centering
    \resizebox{0.35\textwidth}{!}{
        \begin{dependency}
            \begin{deptext}
            영수 \& \textbf{씨는} \& 테니스를 \& 잘 \& 합니다 \\
            \textit{Yengswu} \& \textit{ssi-nun} \& \textit{theynisu-lul} \& \textit{cal} \& \textit{hap-nita}  \\
            Yengswu \& Mr.-TOP \& tennis-ACC \& well \& do-DECL  \\
            \end{deptext}
            \depedge{5}{1}{nsubj}
            \depedge{1}{2}{flat}
            \depedge{5}{3}{obj}
            \depedge{5}{4}{advmod}
            \deproot{5}{root}
        \end{dependency}
    }
    \caption{Flat (Left-headed) \\ \footnotesize{`Youngsoo is good at tennis.'}}
    \label{fig:2}
\end{figure}

\subsubsection{Treatment of auxiliary verbs}

The revised annotation scheme strictly adheres to the UD guidelines for Korean, limiting the annotation of auxiliary verbs to five specific forms.\footnote{\url{https://universaldependencies.org/ko/index.html}} These forms include (1) the affirmative copula 이- (\textit{i-}, "to be"), which is treated as a separate auxiliary even when it functions as a suffix to a nominal predicate;\footnote{When 이- follows a noun and precedes a sentence-final functional morpheme (e.g., -다 \textit{-ta}, as in 친구이다 \textit{chinkwu-i-ta, "is a friend"}), we assigned it the \texttt{root} tag, simplifying the earlier practice of using a special \texttt{root:cop} tag.} (2) the negative copula 않- (\textit{anh-}, "to not be"), annotated as \texttt{AUX} in negative clauses; (3) the affirmative auxiliary 있- (\textit{iss-}, "to be"), used as an auxiliary in affirmative clauses or to indicate progressive aspect; (4) the necessitative modal 하- (\textit{ha-}, "must, should"), which functions as a modal auxiliary expressing necessity; and (5) the desiderative modal 싶- (\textit{sip-}, "will, want"), which serves as a modal auxiliary expressing a desire or intention. Verbs with auxiliary-like meanings outside this set were tagged as adverbial clause modifiers (\texttt{advcl}).

\subsection{Annotation process}

The annotation was conducted by five native Korean speakers, each holding at least an undergraduate degree in Korean linguistics. To manage the workload and ensure comprehensive coverage, the annotators were divided into two groups, with each sentence independently annotated by a pair from one group. The annotators worked independently to minimize bias and preserve the integrity of their individual assessments, without interim adjudication meetings to resolve disagreements. When discrepancies arose between the initial pair of annotators, a third annotator, and if necessary, a fourth, were involved sequentially. Inter-annotator reliability was assessed for the initial annotation pairs (before the adjudication process) using the augmented dataset (\# sents = 4,532, Table \ref{tab:1}).

\begin{table}[h!]
\centering
\resizebox{0.3\textwidth}{!}{%
\begin{tabular}{lr}
Annotation & Cohen's \textit{Kappa} \\
\hline
LEMMA  & 0.964 \\
XPOS   & 0.908 \\
HEAD   & 0.892 \\
DEPREL & 0.927 \\
\end{tabular}
}
\caption{Inter-annotator reliability}
\label{tab:1}
\end{table}

\section{Experiment}
\subsection{Model training}

We evaluated four language models against L2-Korean morphosyntactic annotation tasks, drawing upon user-friendly NLP toolkits designed for multilingual applications in fundamental NLP tasks: (1) \textbf{Baseline:} Stanza-Korean (GSD package) \cite{qi2020stanza} was used as a benchmark without fine-tuning. It aligns with both the Sejong tag set and the UD framework; (2) \textbf{\textit{Stanza}:} We fine-tuned Stanza-Korean (GSD), which employs a biLSTM architecture \cite{huang2015bidirectional} to model sequential dependencies. Fine-tuning allows the model to better capture localized morphosyntactic patterns in L2-Korean data by leveraging the tagging scheme and linguistic patterns encoded in the pre-existing GSD package; (3) \textbf{\textit{spaCy}:} We fine-tuned spaCy \cite{spacy}, which uses its \texttt{tok2vec} layer to generate token-level embeddings from sub-word features. Fine-tuning in spaCy benefits from pre-trained word vectors and built-in lexical resources, making it well-suited for modeling specific lexico-grammatical nuances; (4) \textbf{\textit{Trankit}:} We fine-tuned Trankit \cite{van2021trankit}, which uses a transformer-based architecture (XLM-RoBERTa, \citealp{conneau2020unsupervised}) pre-trained on 100 languages. Fine-tuning a custom pipeline in \textit{Trankit} using the \textit{TPipeline} class enables the model to capture long-range dependencies and complex syntactic structures. All models were trained using their default hyperparameter settings to ensure a fair comparison.

\subsection{Dataset split}

The updated L2-Korean UD treebank (v1.2) was divided into subsets for training and evaluation. The training set contained 9,649 sentences, while the development set, comprising 1,208 sentences, was used for fine-tuning and model optimization. The test set, which included 1,205 sentences, was used to evaluate in-domain performance. Additionally, an out-of-domain test set comprising 922 sentences was designated to assess the models' robustness and generalizability to data beyond the training space.

\subsection{Evaluation Metrics}

To evaluate these models, we measured F1 scores across the following metrics: XPOS, LEMMA, UAS (Unlabeled Attachment Score), and LAS (Labeled Attachment Score).

\subsection{Results}
The fine-tuned models effectively improved their performance across various metrics for both in-domain and out-of-domain datasets. For the in-domain L2K-UD-test set, Trankit outperformed other models in XPOS, UAS, and LAS, while Stanza achieved the best LEMMA score despite trailing overall. In the out-of-domain KoLLA treebank, Trankit again excelled in XPOS, UAS, and LAS, demonstrating its generalizability beyond the traning space. Stanza consistently performed best in the LEMMA metric, indicating its strong lexical capabilities even with domain shifts.
\begin{table}
\centering
\resizebox{0.49\textwidth}{!}{%
\begin{tabular}{crrrrr}
Dataset & Metric & Baseline & Stanza & spaCy & Trankit \\
\hline
\multirow{5}{*}{\shortstack{L2K-UD-test\\(in-domain)}} 
& XPOS    & 82.44    & 89.72   & 83.15 & \textbf{91.81} \\
& LEMMA   & 89.61    & \textbf{95.64} & 87.97 & 88.84\\
& UAS     & 76.72    & 85.53   & 82.21 & \textbf{92.28}\\
& LAS     & 60.69    & 80.36   & 75.21 & \textbf{89.13} \\
\hline
\multirow{5}{*}{\shortstack{KoLLA\\(out-of-domain)}} 
& XPOS    & 77.79    & 81.87   & 71.21 & \textbf{84.51} \\
& LEMMA   & 88.03    & \textbf{91.01}   & 79.64 & 86.90 \\
& UAS     & 72.30    & 81.17   & 74.48 & \textbf{88.93} \\
& LAS     & 58.53    & 75.14   & 63.56 & \textbf{85.45} \\
\end{tabular}
}
\caption{Evaluation metrics}
\label{tab:2}
\end{table}

\section{Discussion and future directions}

We expanded the L2-Korean UD treebank with refined annotation schemes to improve model performance after fine-tuning. Using this treebank, we fine-tuned three models—Stanza, spaCy, and Trankit—and evaluated their performance in both in-domain and out-of-domain contexts. The evaluation results showed significant performance improvements across various metrics, underscoring the value of using an L2 dataset for fine-tuning. Among the models, Trankit’s transformer-based architecture outperformed the others in XPOS, UAS, and LAS across both test datasets, demonstrating its effectiveness of capturing morphosyntactic features in L2-Korean data. The fine-tuned models and relevant documentations are available at \url{https://github.com/NLPxL2Korean/UD-KSL}. The treebank will be updated at \url{https://github.com/UniversalDependencies/UD_Korean-KSL}.

Although both Trankit and Stanza employ a character-based seq2seq model \citep{van2021trankit}, Stanza's superior lemmatization performance compared to Trankit can be attributed to two primary factors. First, Stanza includes a dictionary-based lemmatizer \citep{qi2020stanza}, which may have strengthened its ability to handle a wide variety of morphological patterns. Second, as noted earlier, Stanza uniquely leverages a model that was pre-trained on L1 data (UD-Korean GSD) before being fine-tuned on the current L2 data, which appears to enable it to capitalize on prior lemmatization knowledge for more accurate predictions.

To fully harness the potential of transformer-based architectures in fine-tuning L2-Korean models, future L2-Korean UD treebanks could adopt two complementary strategies. One approach involves combining L2-Korean data drawn from various genres or diverse learner backgrounds. The other centers on refining the match between universal UPOS tags and language-specific XPOS tags through expert revisions to enhance UPOS to boost their effectiveness for lemmatization within the seq2seq framework.

\section*{Acknowledgments}
The authors gratefully acknowledge Hee-June Koh, Chanyoung Lee, Youkyung Sung for their contributions to manual annotations and discussions for the enhancement of annotation guidelines. This research was supported by the 2024 Korean Studies Grant Program of the Academy of Korean Studies (AKS-2024-R-012).

\bibliographystyle{acl_natbib}
\bibliography{nodalida2025}

\newpage

\appendix
\section*{Appendix}

\begin{table}[h!]
\centering
\resizebox{0.49\textwidth}{!}{%
\begin{tabular}{lrrlrr}
\hline
\textbf{XPOS} & \textbf{v1.1} & \textbf{v1.2} & \textbf{DEPREL} & \textbf{v1.1} & \textbf{v1.2} \\ \hline
NNG & 25338 & 40001 & nsubj & 8767 & 13781 \\ \hline
VV & 10219 & 16714 & punct & 8287 & 14066 \\ \hline
EC & 8600 & 13282 & obl & 7332 & 12034 \\ \hline
EF & 7541 & 12994 & root & 6866 & 12989 \\ \hline
SF & 7525 & 12948 & obj & 5572 & 9203 \\ \hline
ETM & 6694 & 9831 & advmod & 4995 & 7829 \\ \hline
JKB & 6366 & 10450 & advcl & 4703 & 8425 \\ \hline
JX & 5406 & 8656 & acl & 4501 & 6400 \\ \hline
NNB & 4748 & 7454 & nmod & 2059 & 3882 \\ \hline
JKO & 4735 & 7717 & aux & 1963 & 2312 \\ \hline
MAG & 4312 & 6774 & conj & 1860 & 2782 \\ \hline
JKS & 4136 & 6668 & amod & 1413 & 2176 \\ \hline
VA & 3380 & 5905 & cc & 1306 & 2154 \\ \hline
XSV & 3278 & 4761 & nmod:poss & 1299 & 1877 \\ \hline
VX & 3237 & 4555 & det & 933 & 1373 \\ \hline
EP & 2850 & 5215 & case & 894 & 1477 \\ \hline
NNP & 2847 & 4810 & flat & 854 & 1172 \\ \hline
NP & 2145 & 3548 & ccomp & 642 & 897 \\ \hline
VCP & 2083 & 3098 & dislocated & 576 & 1035 \\ \hline
MM & 1672 & 2689 & mark & 509 & 838 \\ \hline
XSN & 1467 & 2179 & list & 303 & 444 \\ \hline
JKG & 1329 & 1921 & goeswith & 203 & 235 \\ \hline
NF & 1312 & 2208 & nummod & 179 & 342 \\ \hline
XSA & 1199 & 1815 & appos & 128 & 95 \\ \hline
MAJ & 1160 & 1921 & compound & 52 & 112 \\ \hline
SN & 1017 & 1475 & vocative & 46 & 49 \\ \hline
ETN & 830 & 1213 & parataxis & 37 & 39 \\ \hline
NA & 818 & 1215 & csubj & 22 & 22 \\ \hline
JC & 685 & 1269 & discourse & 6 & 6 \\ \hline
SP & 607 & 864 & fixed & 6 & 24 \\ \hline
XR & 424 & 684 & dep & 3 & 5 \\ \hline
SS & 266 & 378 & &  &  \\ \hline
NV & 262 & 516 & &  &  \\ \hline
VCN & 174 & 251 & &  &  \\ \hline
XPN & 167 & 208 & &  &  \\ \hline
NR & 157 & 228 & &  &  \\ \hline
SL & 133 & 268 & &  &  \\ \hline
JKC & 122 & 177 & &  &  \\ \hline
JKQ & 58 & 86 & &  &  \\ \hline
\end{tabular}
}
\caption{Comparison of XPOS and DEPREL tag distributions in L2-Korean UD v.1.1 and v.1.2}
\label{tab:morpheme_dependency_counts}
\end{table}

\end{document}